# A Flexible Framework for Defeasible Logics

**G. Antoniou** and **D. Billington** and **G. Governatori** and **M.J. Maher**
School of Computing and Information Technology, Griffith University
Nathan, QLD 4111, Australia
{ga,db,guido,mjm}@cit.gu.edu.au

**Abstract**

Logics for knowledge representation suffer from overspecialization: while each logic may provide an ideal representation formalism for some problems, it is less than optimal for others. A solution to this problem is to choose from several logics and, when necessary, combine the representations. In general, such an approach results in a very difficult problem of combination. However, if we can choose the logics from a uniform framework then the problem of combining them is greatly simplified. In this paper, we develop such a framework for defeasible logics. It supports all defeasible logics that satisfy a strong negation principle. We use logic meta-programs as the basis for the framework.

## Introduction

Logics for knowledge representation and, in particular, non-monotonic logics have developed greatly over the past 20 years. Many logics have been proposed, and a deeper understanding of the advantages and disadvantages of particular logics has been developed. There are also, finally, some indications that these logics can be usefully applied (Morgenstern 1998; Prakken 1997).

Unfortunately, it appears that no single logic is appropriate in all situations, or for all purposes. History clearly indicates that while one logic may achieve desired results in some situations, in other situations the outcome is not as successful. This is, no doubt, one reason for the proliferation of non-monotonic logics.

Furthermore, even with a fixed syntax and a common motivating intuition, reasonable people can disagree on the semantics of the logic. This can be seen in the literature on semantics of logic programs with negation, for example, but the point was made more sharply in (Touretzky, Horty and Thomason 1987) where a "clash of intuitions" was demonstrated in several different ways for a simple language describing multiple inheritance with exceptions. So it appears that no single logic, with a fixed semantics, will be appropriate.

However, the diversity of logics threatens to become a Tower of Babel. If different problems require different logics then there are many practical disadvantages: skills in one logic do not transfer to another, combining systems composed of different logics is problematic, etc. It seems unlikely then that these logics are practically useful for knowledge representation.

One way to address this problem is to develop logics that are "tunable" to the situation. That is, to develop a framework of logics in which an appropriate logic can be designed. However, such a framework is not sufficient. Also needed is a methodology for designing logics, and the capability of employing more than one such logic in a representation.

In this paper we develop such a framework for defeasible logics. This is a first step towards addressing the above problem for knowledge representation logics more generally. We make some contributions to the methodology by demonstrating how certain properties can be ensured for a logic. However, there is still much work to be done.

Defeasible logics were introduced and developed by Nute over several years (Nute 1994). These logics perform defeasible reasoning, where a conclusion supported by a rule might be overturned by the effect of another rule. Roughly, a proposition $p$ can be defeasibly proved only when a rule supports it, and it has been demonstrated that no rule supports $\neg p$. These logics also have a monotonic reasoning component, and a priority on rules. One advantage of these logics is that the cost of computing with them is low (Antoniou, Billington, Maher and Rock 2000), in contrast to most logics for knowledge representation.

Nute has developed a framework for defeasible logic that abstracts the many individual logics he has constructed (Nute 1994). Although there are some logics in Nute's framework that cannot be represented in our framework, we will address logics that go well beyond the family of logics addressed by Nute. We consider logics that admit more kinds of conclusions than statements of definite or defeasible proof, as well as logics with different notions of failure-to-prove than the one used in Nute's framework.

In the next section we introduce defeasible logics in general and one particular defeasible logic $DL$. We introduce the Principle of Strong Negation as a design criterion for defeasible logics. In the following sections we demonstrate the framework, first by applying it to $DL$ and then by designing independently motivated variants of $DL$. We also compare it with Nute's framework. In the process, we clarify the relationship between defeasible logics and other non-monotonic logics.

# Defeasible Logics

The family of defeasible logics was introduced by Nute. We begin by outlining the constructs in defeasible logics. We then define the inference rules of a particular defeasible logic $DL$ that has received the most attention. Finally, we introduce the Principle of Strong Negation.

## Outline of Defeasible Logics

A *defeasible theory* $D$ is a triple $(F, R, >)$ where $F$ is a set of literals (called *facts*), $R$ a finite set of rules, and $>$ a superiority relation on $R$. In expressing the proof theory we consider only propositional rules. Rules containing free variables are interpreted as the set of their variable-free instances.

There are three kinds of rules: *Strict rules* are denoted by $A \to p$, and are interpreted in the classical sense: whenever the premises are indisputable (e.g. facts) then so is the conclusion. An example of a strict rule is "Emus are birds". Written formally:

$$emu(X) \to bird(X).$$

Inference from facts and strict rules only is called *definite inference*. Facts and strict rules are intended to define relationships that are definitional in nature. Thus defeasible logics contain no mechanism for resolving inconsistencies in definite inference.

*Defeasible rules* are denoted by $A \Rightarrow p$, and can be defeated by contrary evidence. An example of such a rule is

$$bird(X) \Rightarrow flies(X)$$

which reads as follows: "Birds typically fly".

*Defeaters* are denoted by $A \rightsquigarrow p$ and are used to prevent some conclusions. In other words, they are used to defeat some defeasible rules by producing evidence to the contrary. An example is the rule

$$heavy(X) \rightsquigarrow \neg flies(X)$$

which reads as follows: "If an animal is heavy then it may not be able to fly". The main point is that the information that an animal is heavy is not sufficient evidence to conclude that it doesn't fly. It is only evidence that the animal *may* not be able to fly.

A *superiority relation* on $R$ is an acyclic relation $>$ on $R$ (that is, the transitive closure of $>$ is irreflexive). When $r_1 > r_2$, then $r_1$ is called *superior* to $r_2$, and $r_2$ *inferior* to $r_1$. This expresses that $r_1$ may override $r_2$. For example, given the defeasible rules

$$r : \quad bird(X) \Rightarrow flies(X)$$
$$r' : \quad brokenWing(X) \Rightarrow \neg flies(X)$$

which contradict one another, no conclusive decision can be made about whether a bird with a broken wing can fly. But if we introduce a superiority relation $>$ with $r' > r$, then we can indeed conclude that it cannot fly.

A *conclusion* of a defeasible theory $D$ is a tagged literal. Conventionally (Nute 1994; Billington 1993) there are four tags, so a conclusion has one of the following four forms:

- $+\Delta q$, which is intended to mean that $q$ is definitely provable in $D$.
- $-\Delta q$, which is intended to mean that we have proved that $q$ is not definitely provable in $D$.
- $+\partial q$, which is intended to mean that $q$ is defeasibly provable in $D$.
- $-\partial q$ which is intended to mean that we have proved that $q$ is not defeasibly provable in $D$.

Although the two pairs of tags mentioned above are the only ones currently used in defeasible logics, we will leave open the possibility of further (pairs of) tags. Indeed, we will later introduce in our framework the notion of support for a conclusion, which would require new tags in order to express this notion in a proof theory in the style of the next section.

## Nute's Framework

Nute's framework for defeasible reasoning (Nute 1994) is based around defining a class of proof trees which represent valid inferences. We can reformulate this in terms of conventional inference rules, but we do not have space for a detailed presentation.

Briefly, Nute's framework consists of four inference rules which partly specify the behaviour of the definite (monotonic) reasoning component and its relationship with the defeasible (non-monotonic) reasoning component. Nute defines a defeasible logic to be a logic containing this monotonic kernel of inference rules and satisfying a coherence property. He also discusses several design principles of defeasible logics, but these are not a part of his framework.

## A Defeasible Logic

As an example of a defeasible logic, we consider the logic of (Nute 1987), which we have investigated previously (Antoniou, Billington and Maher 1998; Maher, Antoniu and Billington 1998). In this presentation we use the formulation given in (Billington 1993). We denote this logic by $DL$.

Given a set $R$ of rules, we denote the set of all strict rules in $R$ by $R_s$, the set of strict and defeasible rules in $R$ by $R_{sd}$, the set of defeasible rules in $R$ by $R_d$, and the set of defeaters in $R$ by $R_{dft}$. $R[q]$ denotes the set of rules in $R$ with consequent $q$. In the following $\sim p$ denotes the complement of $p$, that is, $\sim p$ is $\neg p$ if $p$ is an atom, and $\sim p$ is $q$ if $p$ is $\neg q$.

A *rule* $r$ consists of its *antecedent* $A(r)$ (written on the left; $A(r)$ may be omitted if it is the empty set) which is a finite set of literals, an arrow, and its *consequent* $C(r)$ which is a literal. In writing rules we omit set notation for antecedents.

Provability is defined below. It is based on the concept of a *derivation* (or *proof*) in $D = (F, R, >)$. A derivation is a finite sequence $P = (P(1), \ldots P(n))$ of tagged literals satisfying the following conditions. The conditions are essentially inference rules phrased as conditions on proofs. $P(1..i)$ denotes the initial part of the sequence $P$ of length $i$.

$+\Delta$: If $P(i+1) = +\Delta q$ then either
    $q \in F$ or
    $\exists r \in R_s[q]\ \forall a \in A(r) : +\Delta a \in P(1..i)$

$-\Delta$: If $P(i+1) = -\Delta q$ then
    $q \notin F$ and
    $\forall r \in R_s[q]\ \exists a \in A(r) : -\Delta a \in P(1..i)$

$+\partial$: If $P(i+1) = +\partial q$ then either
    (1) $+\Delta q \in P(1..i)$ or
    (2) (2.1) $\exists r \in R_{sd}[q]\ \forall a \in A(r) : +\partial a \in P(1..i)$ and
        (2.2) $-\Delta \sim q \in P(1..i)$ and
        (2.3) $\forall s \in R[\sim q]$ either
            (2.3.1) $\exists a \in A(s) : -\partial a \in P(1..i)$ or
            (2.3.2) $\exists t \in R_{sd}[q]$ such that
                $\forall a \in A(t) : +\partial a \in P(1..i)$ and $t > s$

$-\partial$: If $P(i+1) = -\partial q$ then
    (1) $-\Delta q \in P(1..i)$ and
    (2) (2.1) $\forall r \in R_{sd}[q]\ \exists a \in A(r) : -\partial a \in P(1..i)$ or
        (2.2) $+\Delta \sim q \in P(1..i)$ or
        (2.3) $\exists s \in R[\sim q]$ such that
            (2.3.1) $\forall a \in A(s) : +\partial a \in P(1..i)$ and
            (2.3.2) $\forall t \in R_{sd}[q]$ either
                $\exists a \in A(t) : -\partial a \in P(1..i)$ or $t \not> s$

The elements of a derivation are called *lines* of the derivation. We say that a tagged literal $L$ is *provable* in $D = (F, R, >)$, denoted by $D \vdash L$, iff there is a derivation in $D$ such that $L$ is a line of $P$.

$DL$ is closely related to several non-monotonic logics (Antoniou, Billington and Maher 2000). In particular, the "directly skeptical" semantics of non-monotonic inheritance networks (Horty, Thomason and Touretzky 1987; Horty 1994) can be considered an instance of inference in $DL$ once an appropriate superiority relation, derived from the topology of the network, is fixed (Billington, de Coster, and Nute 1990). $DL$ is a conservative logic, in the sense of Wagner (Wagner 1991).

## The Principle of Strong Negation

The purpose of the $-\Delta$ and $-\partial$ inference rules is to establish that it is not possible to prove a corresponding positive tagged literal. These rules are defined in such a way that all the possibilities for proving $+\partial q$ (for example) are explored and shown to fail before $-\partial q$ can be concluded. Thus conclusions with these tags are the outcome of a constructive proof that the corresponding positive conclusion cannot be obtained.

As a result, there is a close relationship between the inference rules for $+\partial$ and $-\partial$, (and also between those for $+\Delta$ and $-\Delta$). The structure of the inference rules is the same, but the conditions are negated in some sense. We say that the inference rule for $+\partial$ ($-\partial$) is the *strong negation* of the inference rule for $-\partial$ ($+\partial$).

The strong negation of a formula is closely related to the function that simplifies a formula by moving all negations to an innermost position in the resulting formula. It is defined as follows.

$$\begin{aligned}
sneg(+\partial p \in X) &= -\partial p \in X \\
sneg(-\partial p \in X) &= +\partial p \in X \\
sneg(A \wedge B) &= sneg(A) \vee sneg(B) \\
sneg(A \vee B) &= sneg(A) \wedge sneg(B) \\
sneg(\exists x\, A) &= \forall x\, sneg(A) \\
sneg(\forall x\, A) &= \exists x\, sneg(A) \\
sneg(\neg A) &= \neg sneg(A) \\
sneg(A) &= \neg A \quad \text{if } A \text{ is a pure formula}
\end{aligned}$$

A pure formula is a formula that does not contain a tagged literal. Pairs of tags other than $+\partial, -\partial$ are treated in an analogous manner to $+\partial$ and $-\partial$. The strong negation of the applicability condition of an inference rule is a constructive approximation of the conditions where the rule is not applicable.

We are led to consider the following Principle of Strong Negation:

*For each pair of tags such as $+\partial, -\partial$, the inference rule for $-\partial$ should be the strong negation of the inference rule of $+\partial$ (and vice versa).*

Clearly $DL$ satisfies this principle. In fact, all logics in our framework satisfy it. On the other hand, in Nute's framework (Nute 1994) logics may violate it.

There are two other important properties that defeasible logics may have. A theory is *coherent* if there is no $p$ such that $D \vdash +\partial p$ and $D \vdash -\partial p$, or $D \vdash +\Delta p$ and $D \vdash -\Delta p$. A theory is *consistent* if for every $p$ such that $D \vdash +\partial p$ and $D \vdash +\partial \neg p$, also $D \vdash +\Delta p$ and $D \vdash +\Delta \neg p$. Intuitively, coherence says that no literal is simultaneously provable and demonstrably unprovable. Consistency says that a literal and its negation can both be defeasibly provable only when it and its negation are definitely provable; hence defeasible inference does not introduce inconsistency. (As noted earlier, definite provability is intended for definitional information, and has no mechanism for resolving inconsistencies.) A logic is coherent (consistent) if each theory of the logic is coherent (consistent). The above logic $DL$ is coherent and consistent (Billington 1993).

## A Framework of Defeasible Logics

Our framework consists of a meta-program, defining when an atom is definitely or defeasibly proved, and a semantics for the meta-language (which is logic programming). In (Maher and Governatori 1999) it was shown how $DL$ is amenable to definition in this framework. We first introduce the meta-program for $DL$ as a first example of the framework, and then derive some properties of the framework and the logics that can be defined within it. We make a comparison with Nute's framework.

### The $DL$ Meta-program

In this section we introduce a meta-program $\mathcal{M}$ in a logic programming form that expresses the essence of the defeasible reasoning embedded in $DL$. $\mathcal{M}$ consists of the following clauses. We first introduce the predicates defining classes of rules, namely

```
supportive_rule(Name, Head, Body):-
    strict(Name, Head, Body).

supportive_rule(Name, Head, Body):-
    defeasible(Name, Head, Body).

rule(Name, Head, Body):-
    supportive_rule(Name, Head, Body).

rule(Name, Head, Body):-
    defeater(Name, Head, Body).
```

We introduce now the clauses defining the predicates corresponding to $+\Delta$, $-\Delta$, $+\partial$, and $-\partial$. These clauses specify the structure of defeasible reasoning in $DL$. Arguably they convey the conceptual simplicity of $DL$ more clearly than the proof theory.

```
c1    definitely(X):-
          fact(X).

c2    definitely(X):-
          strict(R, X, [Y_1, ..., Y_n]),
          definitely(Y_1),...,definitely(Y_n).

c3    defeasibly(X):-
          definitely(X).

c4    defeasibly(X):-
          not definitely(~X),
          supportive_rule(R, X, [Y_1, ..., Y_n]),
          defeasibly(Y_1),...,defeasibly(Y_n),
          not overruled(R, X).

c5    overruled(R, X):-
          rule(S, ~X, [U_1, ..., U_n]),
          defeasibly(U_1),...,defeasibly(U_n),
          not defeated(S, ~X).

c6    defeated(S, ~X):-
          sup(T, S),
          supportive_rule(T, X, [V_1, ..., V_n]),
          defeasibly(V_1),...,defeasibly(V_n).
```

The first two clauses address definite provability, while the remainder address defeasible provability. The clauses specify if and how a rule in $DL$ can be overridden by another, and which rules can be used to defeat an overriding rule, among other aspects of the structure of defeasible reasoning in $DL$.

We have permitted ourselves some syntactic flexibility in presenting the meta-program. However, there is no technical difficulty in using conventional logic programming syntax to represent this program.

Given a defeasible theory $D = (F, R, >)$, the corresponding program $\mathcal{D}$ is obtained from $\mathcal{M}$ by adding facts according to the following guidelines:

1. `fact(p).` for each $p \in F$
2. `strict(r_i, p, [q_1, ..., q_n]).` for each rule $r_i : q_1, ..., q_n \rightarrow p \in R$
3. `defeasible(r_i, p, [q_1, ..., q_n]).` for each rule $r_i : q_1, ..., q_n \Rightarrow p \in R$
4. `defeater(r_i, p, [q_1, ..., q_n]).` for each rule $r_i : q_1, ..., q_n \leadsto p \in R$
5. `sup(r_i, r_j).` for each pair of rules such that $r_i > r_j$

### The Framework

In previous work (Maher and Governatori 1999) we have established the correctness of this meta-program representation for $DL$. Let $\models_K$ denote logical consequence under Kunen's semantics of logic programs (Kunen 1987).

**Theorem 1** *Let $D$ be a defeasible theory and $\mathcal{D}$ denote its meta-program counterpart.*

*For each literal $p$,*

1. $D \vdash +\Delta p$ iff $\mathcal{D} \models_K$ `definitely`$(p)$;
2. $D \vdash -\Delta p$ iff $\mathcal{D} \models_K \neg$`definitely`$(p)$;
3. $D \vdash +\partial p$ iff $\mathcal{D} \models_K$ `defeasibly`$(p)$;
4. $D \vdash -\partial p$ iff $\mathcal{D} \models_K \neg$`defeasibly`$(p)$;

A significant aspect of this result deserves further comment. Negative conclusions (involving tags $-\Delta$ and $-\partial$), which refer to failure to prove, are characterized by the negation of the positive conclusions. Thus the meta-program implements *failure as negation*.

More generally, this provides a point of comparison between defeasible logics and other non-monotonic logics: in defeasible logics failure is the basic notion, whereas negation is basic in most other non-monotonic logics. Nevertheless, these two notions are different sides of the same coin.

An important feature of the meta-programming framework for defeasible logic is that it admits different forms of failure, corresponding to different semantics of negation in logic programs. For example, under the Kunen semantics (and hence in $DL$) with the program

$p \Rightarrow p$

we cannot conclude $-\partial p$, even though it is clear that $+\partial p$ will never be concluded. Under the well-founded semantics (Van Gelder, Ross and Schlipf 1991) we can conclude $-\partial p$. (Maher and Governatori 1999) contains the definition of a defeasible logic that results from using the well-founded semantics with the above meta-program $\mathcal{M}$.

*Our framework consists of a meta-program defining* `defeasibly` *and* `definitely`, *among other predicates, the implicit definition of negative tags by the negation of these predicates, and a semantics for the meta-language (logic programming).*

Every logic defined within the framework satisfies the Principle of Strong Negation, by construction. We say that a semantics for logic programs is *consistent* if for no program $P$ and atom $a$ does the semantics of $P$ imply both $a$ and $\neg a$ are true. Thus

**Theorem 2** *Every defeasible logic defined in our framework using a consistent semantics is coherent.*

Furthermore, containment relations among different positively tagged conclusions are reflected in the corresponding negatively tagged conclusions. For any tag $\lambda$ and logic $L$, we also use $\lambda$ to denote the set of literals $p$ such that $\lambda p$ can be concluded by $L$.

**Proposition 3** *Let $+\tau_1, -\tau_1$ and $+\tau_2, -\tau_2$ be pairs of tags in a logic defined in our framework. Then $+\tau_1 \subseteq +\tau_2$ iff $-\tau_1 \supseteq -\tau_2$.*

We can characterize the extent to which Nute's framework is covered by ours.

**Theorem 4** *Every defeasible logic in Nute's framework that satisfies the Principle of Strong Negation can be represented in our framework, using Kunen's semantics.*

The proof is based on the form of permitted inference rules (Nute 1994) and the transformation of arbitrary logic formulas to logic programs used in (Lloyd and Topor 1984).

In view of this result and the consistency of Kunen's semantics we can establish that all such logics are coherent.

**Corollary 5** *Every defeasible logic in Nute's framework that satisfies the Principle of Strong Negation is coherent.*

The presence of the Kunen semantics provides substantial insight into the computational complexity of defeasible logics. It means that every defeasible logic in Nute's sense that admits free variables and function symbols, and satisfies the Principle of Strong Negation is computable, in contrast to the great majority of non-monotonic logics which are uncomputable. Similarly, if we consider only propositional logics then, under certain restrictions on the meta-program, the consequences of a theory can be computed in polynomial time[1]. Again, this is in contrast to the great majority of non-monotonic logics.

There are several points of difference between our framework and Nute's.

- Nute's framework is committed to a very specific (though natural) notion of failure-to-prove: the one corresponding to the Kunen semantics. Our framework is not restricted in this way.
- Nute's framework is able to express logics that violate the Principle of Strong Negation, whereas ours cannot.
- By admitting arbitrary inference rules (in addition to the monotonic kernel) but requiring coherence, Nute's framework places the burden of proof that the result is a defeasible logic on the logic designer. Every logic designed within our framework is coherent.
- The setting of Nute's framework makes it extremely difficult to handle defeasible rules containing free variables and function symbols. These can be handled very naturally in the meta-programming framework.
- It is not clear whether the four tags are intended to be the only tags admissible in Nute's framework or not. In the following section, we will demonstrate the advantage of admitting other tags.
- There are some technical differences between Nute's framework and $DL$, concerning the relative priority of strict and defeasible rules in cases where antecedents are known defeasibly. These arise from Nute's monotonic kernel, which has some differences with the inference rules in $DL$.

---

[1] Indeed, $DL$ has been shown to have linear complexity (Antoniou, Billington, Maher and Rock 2000).

## New Defeasible Logics

We now develop several variations of $DL$. Our interest here is not to develop definitive defeasible logics, but to demonstrate the flexibility of the framework, and the beginnings of a methodology for designing logics. We have already, in (Maher and Governatori 1999), defined an extension of $DL$ to allow a failure operator in the body of rules without disturbing the semantics of $DL$ on theories without this operator. To keep this paper brief, we ignore definite inference in this section. A key element of the definition of the logics is the notion of support, used as part of Wagner's analysis of defeasible reasoning (Wagner 1991), so we begin by finding an appropriate definition of support.

### Support

Support for a literal $p$ consists of a chain of reasoning that would lead us to conclude $p$ in the absence of conflicts. If we ignore the superiority relation we could define it simply as follows.

$c7$    `supported`$(X)$:-
      `definitely`$(X)$.

$c8$    `supported`$(X)$:-
      `supportive_rule`$(R, X, [Y_1, \ldots, Y_n])$,
      `supported`$(Y_1)$,…,`supported`$(Y_n)$.

However, in situations where two conflicting rules can be applied and one rule is inferior to another, the inferior rule should not be counted as supporting its conclusion. Thus we refine $c8$:

$c9$    `supported`$(X)$:-
      `supportive_rule`$(R, X, [Y_1, \ldots, Y_n])$,
      `supported`$(Y_1)$,…,`supported`$(Y_n)$,
      `not beaten`$(R, X)$.

$c10$    `beaten`$(R, X)$:-
      `rule`$(S, \sim X, [W_1, \ldots, W_n])$,
      `defeasibly`$(W_1)$,…,`defeasibly`$(W_n)$,
      `sup`$(S, R)$.

Notice that, because the definition of `support` is recursive, we would not be able to express it in the proof theories of (Nute 1994; Billington 1993) without additional tags.

### Ambiguity Propagation

A literal is *ambiguous* if there is a chain of reasoning that supports a conclusion that $p$ is true, another that supports that $\neg p$ is true, and the superiority relation does not resolve this conflict.

**Example 1** The following is a classic example of non-monotonic inheritance.

   $r_1 : \Rightarrow quaker$
   $r_2 : \Rightarrow republican$
   $r_3 : quaker \Rightarrow pacifist$
   $r_4 : republican \Rightarrow \neg pacifist$
   $r_5 : republican \Rightarrow footballfan$
   $r_6 : pacifist \Rightarrow antimilitary$
   $r_7 : footballfan \Rightarrow \neg antimilitary$

The priority relation is empty.

$pacifist$ is ambiguous since the combination of $r_1$ and $r_3$ support $pacifist$ and the combination of $r_2$ and $r_4$ support $\neg pacifist$. Similarly, $antimilitary$ is ambiguous.

In $DL$, the ambiguity of $pacifist$ results in the conclusions $-\partial pacifist$ and $-\partial \neg pacifist$. Since $r_6$ is consequently not applicable, $DL$ concludes $+\partial \neg antimilitary$. This behaviour is called *ambiguity blocking*, since the ambiguity of $antimilitary$ has been blocked by the conclusion $-\partial pacifist$ and an unambiguous conclusion about $antimilitary$ has been drawn.

A preference for ambiguity blocking or ambiguity propagating behaviour is one of the properties of non-monotonic inheritance nets over which intuitions can clash (Touretzky, Horty and Thomason 1987). Stein (Stein 1992) argues that ambiguity blocking results in an unnatural pattern of conclusions in extensions of the above example. Ambiguity propagation results in fewer conclusions being drawn, which might make it preferable when the cost of an incorrect conclusion is high. For these reasons an ambiguity propagating variant of $DL$ is of interest.

We can achieve ambiguity propagation behaviour by making a minor change to clause $c5$ so that it now considers support to be sufficient to allow a superior rule to overrule an inferior rule.

$c11$   `overruled(R, X):-`
        `rule(S, ~ X, [U_1, ..., U_n]),`
        `supported(U_1),...,supported(U_n),`
        `not defeated(S, ~ X).`

**Proposition 6** *The resulting logic is consistent.*

Applying this logic to the example above, all literals mentioned in the theory (both positive and negated) are supported. As in $DL$, we conclude $-\partial pacifist$ and $-\partial \neg pacifist$, since $r_3$ and $r_4$ overrule each other. We also conclude $+\partial footballfan$ and $-\partial antimilitary$ for essentially the same reason as in $DL$. However this logic differs from $DL$ and propagates ambiguity by concluding $-\partial \neg antimilitary$, since $r_7$ is overruled by $r_6$ and $r_7$ cannot defeat $r_6$.

### Team Defeat

The defeasible logics we have considered so far incorporate the idea of *team defeat*. That is, an attack on a rule with head $p$ by a rule with head $\sim p$ may be defeated by a *different* rule with head $p$ (see inference rule $+\partial$ and clauses $c5$ and $c6$). Even though the idea of team defeat is natural, it is worth noting that several related approaches, such as LPwNF (Dimopoulos and Kakas 1995) and most argumentation frameworks, do not adopt this idea. It is easy to define defeasible logics without team defeat in our framework. For our original defeasible logic ($c1$–$c6$) this can be achieved by replacing $c5$ and $c6$ by the following clause.

$c12$   `overruled(R, X):-`
        `rule(S, ~ X, [U_1, ..., U_n]),`
        `defeasibly(U_1),...,defeasibly(U_n),`
        `not sup(R, S).`

**Proposition 7** *The resulting logic is consistent.*

It is important to note that the two techniques demonstrated here are orthogonal, in the sense that they can be applied individually or in combination, without prejudice. For example, we may define an ambiguity propagating defeasible logic without team defeat. This is achieved by replacing the clauses $c11$ and $c6$ by the following new one:

$c13$   `overruled(R, X):-`
        `rule(S, ~ X, [U_1, ..., U_n]),`
        `supported(U_1),...,supported(U_n),`
        `not sup(R, S).`

**Proposition 8** *The resulting logic is consistent.*

In this sense we have established a tunable framework in which a defeasible logic may be designed according to the specific needs of the problem at hand.

### Relationships

In this section we wish to establish relationships among some of the variants we introduced in this paper. We will show that there exists a chain of increasing expressive power among several of the logics. We will be considering the following tags:

- $\Delta$, which denotes strict provability ($c1, c2$).
- $\partial_{a,ntd}$, which denotes defeasible provability in the ambiguity propagating logic without team defeat ($c1$–$c4, c13$).
- $\partial_a$, which denotes defeasible provability in the the ambiguity propagating defeasible logic ($c1$–$c4, c7, c9, c10, c11, c6$).
- $\partial$, which denotes defeasible provability in our original defeasible logic ($c1$–$c6$).
- $\Sigma$, which denotes support in our original defeasible logic ($c7, c9, c10$).

Then we are able to prove the following:

**Theorem 9** $+\Delta \subset +\partial_{a,ntd} \subset +\partial_a \subset +\partial \subset +\Sigma$.
*Each inclusion is strict, in the sense that there are defeasible theories in which the inclusion is strict.*

We wish to point out that this result is deeper that it may look on the surface. For example, the relation $+\partial_{a,ntd} \subset +\partial_a$ appears trivial since the absence of team defeat makes the logic weaker. But notice that when the logic fails to prove a literal $p$ and instead proves $-\partial p$, then that result may be used by the logic to prove another literal $q$ that could not be proven if $p$ were provable. In fact it is easily seen that defeasible provability in the original defeasible logic without team defeat is *not* weaker than defeasible provability with team defeat. Consider the following example:

**Example 2**

$r_1 :\Rightarrow p$
$r_2 :\Rightarrow p$
$r_3 :\Rightarrow \neg p$
$r_4 :\Rightarrow \neg p$
$r_5 : p \Rightarrow \neg q$
$r_6 :\Rightarrow q$
$r_1 > r_3, \; r_2 > r_4$

Then $q$ is not defeasibly provable in the original defeasible logic, but defeasibly provable in the logic without team defeat.

## Conclusion

We have developed a framework for defeasible logics that admits a wide range of logics. We have demonstrated the flexibility of the framework and the beginnings of a design methodology by developing, in a straightforward way, variants of $DL$ which are, respectively, ambiguity propagating and incapable of team defeat. All logics designed within the framework are coherent.

The uniform setting provided by logic meta-programming supports the easy combination of logics that are based on the same form of failure. We have a proposal for combining logics with different notions of failure, based on the module system of (Maher 1993), but we have no space to present it here.

In summary, our framework provides a tunable family of defeasible logics.

## Acknowledgements

This research was supported by the Australia Research Council. under Large Grant No. A49803544.

## References

[Antoniou, Billington and Maher 1998] G. Antoniou, D. Billington and M.J. Maher. 1998. Normal Forms for Defeasible Logic. In *Proc. Joint International Conference and Symposium on Logic Programming*, J. Jaffar (Ed.), 160–174. MIT Press, 1998.

[Antoniou, Billington, Maher and Rock 2000] G. Antoniou, D. Billington, M.J. Maher, A. Rock. 2000. Efficient Defeasible Reasoning Systems, *Proc. Australian Workshop on Computational Logic*.

[Antoniou, Billington and Maher 2000] G. Antoniou, M.J. Maher, and D. Billington. 2000. Defeasible Logic versus Logic Programming without Negation as Failure. *Journal of Logic Programming* (2000).

[Billington, de Coster, and Nute 1990] D. Billington, K. de Coster and D. Nute. 1990. A Modular Translation from Defeasible Nets to Defeasible Logic. *Journal of Experimental and Theoretical Artificial Intelligence* 2: 151–177.

[Billington 1993] D. Billington. 1993. Defeasible Logic is Stable. *Journal of Logic and Computation* 3: 370–400.

[Dimopoulos and Kakas 1995] Y. Dimopoulos and A. Kakas. 1995. Logic Programming without Negation as Failure. In *Proc. ICLP-95*, MIT Press.

[Horty 1994] J.F. Horty. Some Direct Theories of Nonmonotonic Inheritance. In D.M. Gabbay, C.J. Hogger and J.A. Robinson (eds.): *Handbook of Logic in Artificial Intelligence and Logic Programming Vol. 3*, 111–187, Oxford University Press, 1994.

[Horty, Thomason and Touretzky 1987] J.F. Horty, R.H. Thomason and D. Touretzky. 1987. A Skeptical Theory of Inheritance in Nonmonotonic Semantic Networks. In *Proc. AAAI-87*, 358–363.

[Kunen 1987] K. Kunen. 1987. Negation in Logic Programming. *Journal of Logic Programming* 4: 289–308.

[Lloyd and Topor 1984] J. W. Lloyd and R. W. Topor. 1984. Making Prolog more Expressive. *Journal of Logic Programming* 1(3): 225–240.

[Maher 1993] M.J. Maher. 1993. A Transformation System for Deductive Database Modules with Perfect Model Semantics. *Theoretical Computer Science 110,* 377–403.

[Maher, Antoniu and Billington 1998] M. Maher, G. Antoniou and D. Billington. 1998. A Study of Provability in Defeasible Logic. In *Proc. Australian Joint Conference on Artificial Intelligence*, 215–226, LNAI 1502, Springer.

[Maher and Governatori 1999] M. Maher and G. Governatori. 1999. A Semantic Decomposition of Defeasible Logics. *Proc. American National Conference on Artificial Intelligence (AAAI-99),* 299–306.

[Morgenstern 1998] L. Morgenstern. 1998. Inheritance Comes of Age: Applying Nonmonotonic Techniques to Problems in Industry. *Artificial Intelligence*, 103, 1–34.

[Nute 1987] D. Nute. 1987. Defeasible Reasoning. In *Proc. 20th Hawaii International Conference on Systems Science*, IEEE Press, 470–477.

[Nute 1994] D. Nute. 1994. Defeasible Logic. In D.M. Gabbay, C.J. Hogger and J.A. Robinson (eds.): *Handbook of Logic in Artificial Intelligence and Logic Programming Vol. 3*, Oxford University Press, 353–395.

[Prakken 1997] H. Prakken. 1997. *Logical Tools for Modelling Legal Argument: A Study of Defeasible Reasoning in Law.* Kluwer Academic Publishers.

[Stein 1992] L.A. Stein. 1992. Resolving Ambiguity in Nonmonotonic Inheritance Hierarchies. *Artificial Intelligence* 55: 259–310.

[Touretzky, Horty and Thomason 1987] D.D. Touretzky, J.F. Horty and R.H. Thomason. 1987. A Clash of Intuitions: The Current State of Nonmonotonic Multiple Inheritance Systems. In *Proc. IJCAI-87*, 476–482, Morgan Kaufmann, 1987.

[Van Gelder, Ross and Schlipf 1991] A. Van Gelder, K. Ross and J.S. Schlipf. Unfounded Sets and Well-Founded Semantics for General Logic Programs. *Journal of the ACM* 38 (1991): 620–650.

[Wagner 1991] G. Wagner. 1991. Ex Contradictione Nihil Sequitur. In *Proc. IJCAI-91*, 538–546, Morgan Kaufmann.